\documentclass[letterpaper, 10 pt, conference]{ieeeconf}  

\IEEEoverridecommandlockouts                              
\usepackage[utf8]{inputenc}
\usepackage{footmisc}
 \usepackage[table]{xcolor} 
\usepackage{booktabs} 
\let\labelindent\relax
\usepackage{enumitem}
\usepackage{cite}
\usepackage{graphicx}
\usepackage{textcomp}
\usepackage{amsmath,amssymb,amsfonts}

\usepackage{algorithm}
\usepackage{algorithmicx}
\usepackage[noend]{algpseudocode}
\usepackage{listings}
\usepackage{xcolor}

\usepackage[hyphens]{url}
\usepackage[hidelinks]{hyperref}
 
\usepackage{cleveref}
\usepackage[makeroom]{cancel}
\usepackage{xcolor,colortbl}
\def\BibTeX{{\rm B\kern-.05em{\sc i\kern-.025em b}\kern-.08em
    T\kern-.1667em\lower.7ex\hbox{E}\kern-.125emX}}

\newcommand{\ourmodel}{COLLAGE\xspace}

\usepackage{tikz}
\usetikzlibrary{positioning}
\usepackage{multirow,booktabs,multicol}

\usepackage{url}

\newcommand\hmm[1]{\ifnum\ifhmode\spacefactor\else2000\fi>1000 \uppercase{#1}\else#1\fi}

\usepackage{setspace}
\usepackage{xspace}
\usepackage[export]{adjustbox}

\usepackage{enumitem}

\algnewcommand{\algorithmicgoto}{\textbf{go to}}%
\algnewcommand{\Goto}[1]{\algorithmicgoto~\ref{#1}}%

\makeatletter
\newcommand*{\rom}[1]{\expandafter\@slowromancap\romannumeral #1@}
\makeatother

\title{\LARGE \textbf{COLLAGE}: Collaborative Human-Agent Interaction Generation using Hierarchical Latent Diffusion and Language Models\vspace{-2mm}}

 \author{Divyanshu Daiya$^{1}$, Damon Conover$^{2}$,  Aniket Bera$^{1}$\\
$^1$Department of Computer Science, Purdue University  $^2$DEVCOM Army Research Laboratory\\
\texttt{\{divyanshu, aniketbera\}@purdue.edu, damon.m.conover.civ@army.mil}
}

\begin{document}

\maketitle
\thispagestyle{empty}
\pagestyle{empty}
\begin{abstract}
We propose a novel framework COLLAGE for generating collaborative agent-object-agent interactions by leveraging large language models (LLMs) and hierarchical motion-specific vector-quantized variational autoencoders (VQ-VAEs). Our model addresses the lack of rich datasets in this domain by incorporating the knowledge and reasoning abilities of LLMs to guide a generative diffusion model. The hierarchical VQ-VAE architecture captures different motion-specific characteristics at multiple levels of abstraction, avoiding redundant concepts and enabling efficient multi-resolution representation. We introduce a diffusion model that operates in the latent space and incorporates LLM-generated motion planning cues to guide the denoising process, resulting in prompt-specific motion generation with greater control and diversity. Experimental results on the CORE-4D, and InterHuman datasets demonstrate the effectiveness of our approach in generating realistic and diverse collaborative human-object-human interactions, outperforming state-of-the-art methods. Our work opens up new possibilities for modeling complex interactions in various domains, such as robotics, graphics and computer vision. 
\end{abstract}

\section{Introduction}
\label{sec:intro}
Modeling human-like agent-object interactions is fundamental in the vision community, enabling applications in gaming, embodied AI, robotics, and VR/AR. While recent works have explored single-person and multi-human object interactions in non-collaborative settings \cite{CG-HOI,CHOIS,HOI-Diff,wu2024thor,InterDreamer,zhang2024hoi}, generating collaborative human-object-human interactions remains largely unexplored. This task requires a complex understanding of human actions and object interactions, as guiding individual agents along with the task involves extensive planning. Given the lack of rich datasets, training a generalized model is challenging. To address this, we propose incorporating the knowledge and reasoning abilities of large language models (LLMs) to guide a generative diffusion latent diffusion model for multi-human-object motion generation in collaborative settings. In the remainder of this paper, we will use the terms `human' and `agent' interchangeably, with the specific application determining the appropriate usage. For robotics applications, `agent' may refer to either a real human or a robotic, human-like entity such as a humanoid.

\begin{figure}
    \centering
\includegraphics[width=0.85\linewidth]{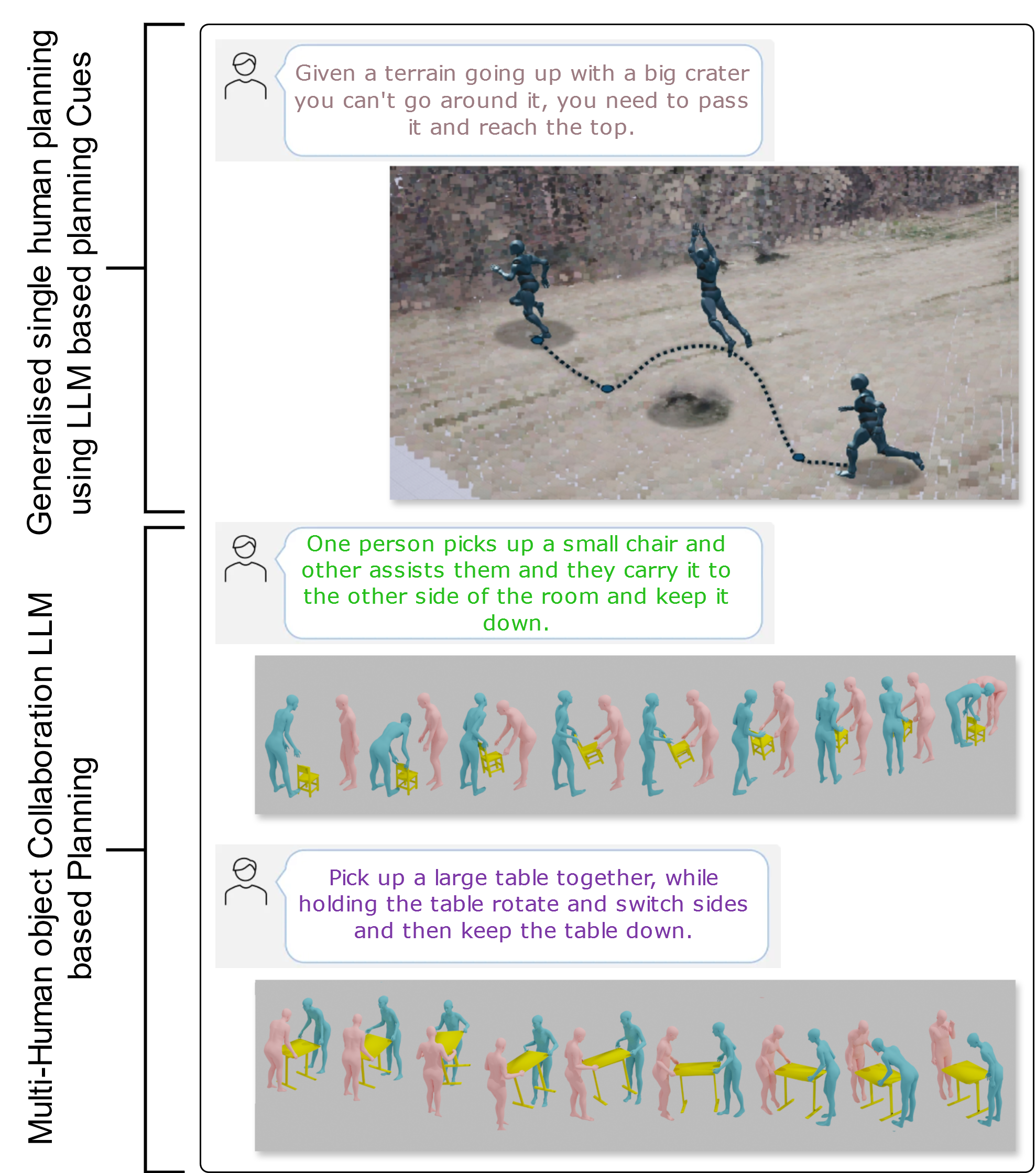}
    \caption{ \textit{Text to collaborative motion and generalized motion generation by \ourmodel, based on user-provided text prompts. In the top image, a simulated humanoid robot adapts to the 3D terrain features based on the input text from the human collaborator. In the bottom image, the two human agents collaborate to handle an object using LLM-based planning via our architecture.}}
    \vspace{-7mm}
    \label{fig:demo}
\end{figure}
Pre-trained LLMs, such as GPT-4 \cite{chatgpt} and Llama 2 \cite{touvron2023llama}, have demonstrated emergent capabilities in reasoning, planning, and motion planning \cite{kojima2022large,huang2022language,huang2023inner,sun2024prompt}. We hypothesize that LLMs could provide a general and domain-independent approach to modeling and planning interactive multi-human object and human-object-human task collaboration, given proper learning approaches. Learning to plan without a dataset can help with motion planning in outdoor settings, where currently, no dataset exists with extensive motion capture data. Utilizing humanoid robots in such settings is a significant hurdle, and effective use of planning via LLMs for fine-grained motion generation could help with humanoid-based motion and interaction in outdoor environments\footnote{Dataset from  GDIS Terrain Segmentation Dataset (Havre de Grace, MD), curated by DEVCOM Army Research Lab} (Fig. \ref{fig:demo}).
To capture the complex motion dynamics in collaborative settings, we propose a hierarchical motion-specific vector-quantized variational autoencoder (VQ-VAE) architecture that explicitly captures different motion-specific characteristics at different levels of abstraction, addressing the limitations of previous VQ-VAE models \cite{zhang2023t2m,kong2023priority,li2021task}.
We incorporate a diffusion model \cite{kong2023priority,Kong23m2dm,chen2023mld} for learning human motion in the latent space and propose a novel architecture to incorporate multi-human object interactions. We augment textual planning cues from LLM with codebook-based associations learned via VQ-VAE training, helping the diffusion model better learn to model according to the task description, given the complex interaction setting and associations. We showcase how plans and cues generated by LLMs can be utilized by the diffusion model for effective generation, demonstrating faster speed and greater generational diversity compared to other models.

We evaluate our model's generalizability on the human-object-human CORE-4D dataset \cite{zhang2024core4d} and multi-human dataset like InterGen \cite{liang2023intergen}, comparing our results with existing works in this setting.


The main contributions of our work are:
\begin{itemize}
\item We propose a novel approach for generating collaborative human-object-human interactions by leveraging LLMs and hierarchical motion-specific VQ-VAEs, addressing the lack of rich datasets in this domain.
\item We introduce a hierarchical motion-specific VQ-VAE architecture that captures different motion-specific characteristics at different levels of abstraction, avoiding redundant repeated concepts across layers and enabling efficient multi-resolution representation.
\item We demonstrate the effectiveness of utilizing LLM-generated motion planning cues to guide the diffusion model through the denoising process, resulting in prompt-specific motion generation with greater control and diversity.
\item We evaluate our model on multiple datasets, showcasing its generalizability and effectiveness in generating collaborative human-object-human interactions.
\end{itemize}
\section{Related Work}
\label{sec:related}
\noindent\textbf{Text-Conditioned Human Motion Generation.}
Generating human motions based on textual descriptions has been a recent research focus. Early approaches generated motions based on action categories~\cite{guo2020action2motion,petrovich2021action,lee2023multiact,athanasiou2022teach}, past motions~\cite{yuan2020dlow,mao2021generating,barquero2022belfusion,chen2023humanmac,xu22stars,xu2023stochastic}, trajectories~\cite{kaufmann2020convolutional,karunratanakul2023gmd,rempe2023trace,xie2023omnicontrol,wan2023tlcontrol}, and scene context~\cite{cao2020long,hassan2021populating,wang2021synthesizing,wang2021scene,wang2022towards,wang2022humanise,huang2023diffusion,zhao2022compositional,Zhao:ICCV:2023,tendulkar2022flex,zhang2023roam}. Recent works have enabled direct generation of human motions from textual inputs~\cite{petrovich2023tmr,guo2022tm2t,petrovich22temos,chen2023executing,zhang2022motiondiffuse,zhang2023motiongpt,zhang2023generating,tevet2022motionclip,ahuja2019language2pose,guo2022generating,kim2023flame,lu2023humantomato,raab2023single,yonatan2023,dabral2023mofusion,wei2023understanding,zhang2023tedi,kong2023priority,yazdian2023motionscript,barquero2024seamless,zhou2023emdm,ma2024contact}, extending to multi-person~\cite{liu2023interactive,wang2023intercontrol,ghosh2023remos} and human-scene interactions~\cite{huang2023diffusion,jiang2024scaling,cong2024laserhuman}. However, generating collaborative human-object-human interactions remains largely unexplored.\\
\noindent\textbf{Human-Object Interaction Generation.}
Modeling realistic human-object interactions is challenging due to the complexity of capturing both human motions and object dynamics. Prior research has addressed hand-object interactions~\cite{li2023task,ye2023affordance,zheng2023cams,zhou2022toch,zhang2023artigrasp}, single-frame human-object interactions~\cite{xie2022chore,zhang2020perceiving,wang2022reconstructing,petrov2023object,hou2023compositional,kim2023ncho}, and zero-shot settings~\cite{li2024genzi,yang2023lemon,kim2024zero}. Recent studies have explored whole-body dynamic interaction generation through kinematic-based~\cite{starke2019neural,starke2020local,taheri2022goal,wu2022saga,kulkarni2023nifty,zhang2022couch,lee2023locomotion,xu2021d3dhoi,ghosh2022imos,corona2020context,9714029,razali2023action,Mandery2015a,Mandery2016b,krebs2021kit,xu2023interdiff,li2023object,wu2024thor} and physics-based methods~\cite{liu2018learning,chao2021learning,merel2020catch,hassan2023synthesizing,bae2023pmp,yang2022learning,xie2022learning,xie2023hierarchical,pan2023synthesizing,braun2023physically,wang2023physhoi,cui2024anyskill}, but often suffer from limitations such as a narrow scope of actions, static objects, or lack of comprehensive whole-body motion representation.\\
\noindent\textbf{Collaborative Multi-Human Interaction Modeling.}
Collaborative human-object-human interactions remain largely unexplored, despite the study of multi-human interactions in non-collaborative contexts~\cite{zhang2024hoi}. The complexity arises from modeling intricate coordination between multiple humans and objects, requiring advanced planning and understanding of collective actions. Recent datasets and baselines, such as CORE-4D~\cite{zhang2024core4d}, have begun to address this gap, but further research is needed to develop models capable of handling such complex interactions.\\
\noindent\textbf{Utilizing LLMs in Motion Generation.}
Large language models (LLMs) have demonstrated remarkable abilities in reasoning~\cite{kojima2022large}, planning~\cite{huang2022language}, and task execution~\cite{huang2023inner}. In the realm of digital humans, LLMs have been employed to guide motion generation~\cite{athanasiou2023sinc,yao2023moconvq,jiang2023motiongpt,zhang2023motiongpt,xiao2023unified}. Our approach extends this line of work by utilizing LLMs to guide the generation of collaborative human-object-human interactions.\\
\noindent\textbf{Hierarchical VQ-VAE and Diffusion Models in Motion Generation.}
Vector-Quantized Variational Autoencoders (VQ-VAEs) have been used to create quantized motion latent spaces~\cite{zhang2023t2m,kong2023priority,jiang2023motiongpt}, but may struggle with complex and diverse motion generation due to limitations like small codebook, as increasing the codebook size for incorporating the complex dataset leads to issue of codebook collapse, as we observed while generalizing ~\cite{kong2023priority} for multi human setting, further smaller codebooks even in single human setting results in less diverse motion. Hierarchical architectures have been proposed to enhance motion modeling~\cite{li2021task}. Diffusion models have also been employed for learning human motion in latent spaces~\cite{kong2023priority,Kong23m2dm,chen2023mld}. Our approach incorporates a hierarchical motion-specific VQ-VAE architecture and a diffusion model guided by LLM-generated plans to effectively generate collaborative human-object-human interactions.
\section{Methodology}
\subsection{Hierarchical VQ-VAE with Description Cues}
Modeling complex human-object interactions necessitates capturing motion dynamics at multiple levels of abstraction, from high-level trajectories and interaction types to low-level limb movements and object manipulations. To achieve this, we propose a hierarchical Vector Quantized Variational Autoencoder (VQ-VAE) that incorporates description cues provided by a Language Model (LLM) at each level of abstraction. This architecture enables the model to learn disentangled motion representations corresponding to different semantic concepts guided by hierarchical textual cues.

\begin{figure*}[!t] 
    \centering
    \includegraphics[width=\linewidth]{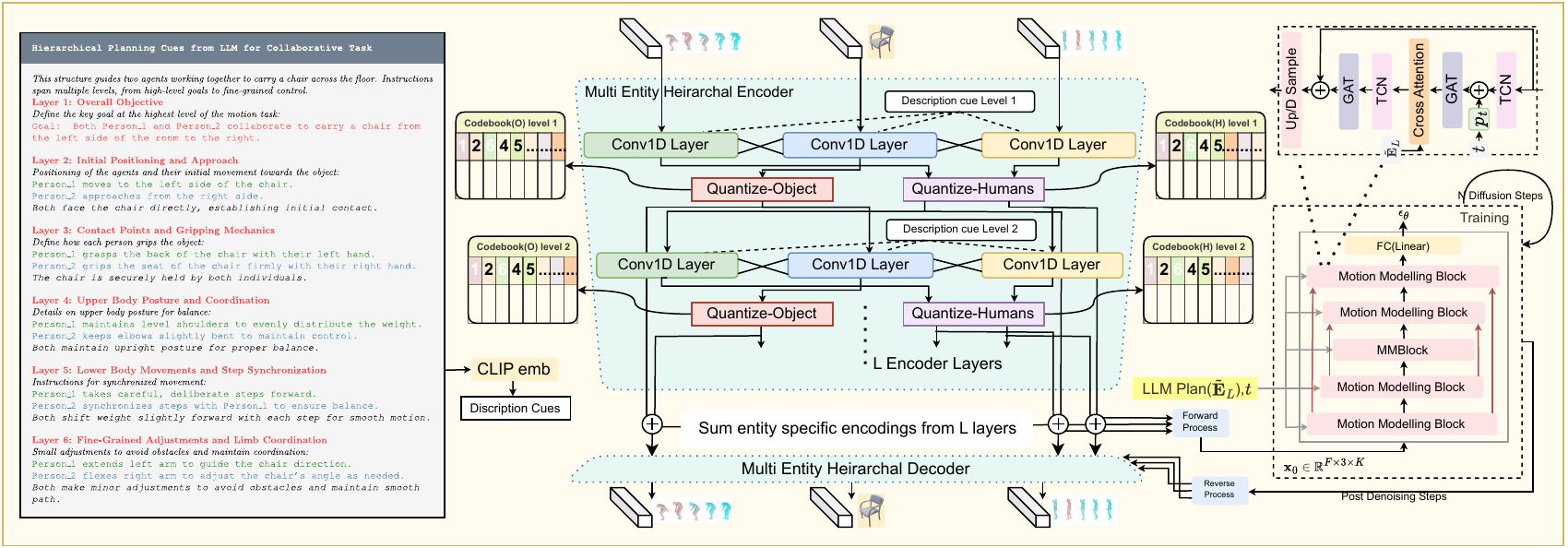}
    \vspace{-5mm} 
    \caption{ \textit{Overview of the proposed \textbf{COLLAGE} framework for collaborative human-object interaction generation. The hierarchical VQ-VAE encoder captures motion-specific characteristics at different levels of abstraction. The latent diffusion model operates in the learned latent space and incorporates LLM-generated motion planning cues to guide the denoising process, enabling the generation of prompt-specific interactions with enhanced control and diversity as in Fig~\ref{fig:demo}.}}
    \vspace{-6mm} 
    \label{fig:pipeline}
\end{figure*}

Our hierarchical VQ-VAE architecture captures motion dynamics at multiple levels of abstraction, as shown in \ref{fig:pipeline}. The encoders at each level map the inputs to latent representations, which are then quantized using codebooks. The decoders reconstruct the original data from the quantized latent representations.
At each level $l$, the encoder for human $i$ computes $Z_H^{i,(l)} = E_H^{(l)}\left( Z_H^{i,(l-1)}; \theta_H^{(l)} \right)$, where $E_H^{(l)}$ is a neural network with parameters $\theta_H^{(l)}$, and $Z_H^{i,(0)} = X^i$ is the input sequence for human $i$. Similarly, the object encoder computes $Z_O^{(l)} = E_O^{(l)}\left( Z_O^{(l-1)}; \theta_O^{(l)} \right)$, with $Z_O^{(0)} = Y$.
We incorporate description cues $\mathbf{e}^{(l)}$ provided by an LLM at each level $l$, which are integrated into the encoder by augmenting the latent representations:
$\tilde{Z}_H^{i,(l)} = \text{Concat}\left( Z_H^{i,(l)}, \mathbf{e}_H^{(l)} \right), \tilde{Z}_O^{(l)} = \text{Concat}\left( Z_O^{(l)}, \mathbf{e}_O^{(l)} \right)$,
where $\mathbf{e}_H^{(l)}$ and $\mathbf{e}_O^{(l)}$ are the description embeddings for humans and objects at level $l$, respectively.

Multi-head attention mechanisms are employed to capture interactions between all pairs of entities, including \( n \) humans and m objects. We compute the attention for each entity \( i \) at level \( l \) across all pairs involving entity \( i \), $ A_{total}^{i,(l)} = \sum_{j \neq i} \text{MultiHeadAttention}\left( Q^{i,(l)}, K^{j,(l)}, V^{j,(l)} \right)$, where \( Q^{i,(l)} \), \( K^{j,(l)} \), and \( V^{j,(l)} \) are the query, key, and value matrices respectively, and the summation extends over all other entities \( j \). The updated latent vector for entity \( i \) after applying attention and layer normalization, $
\hat{Z}_H^{i,(l)} = \text{LayerNorm}\left( \tilde{Z}_H^{i,(l)} + A_{total}^{i,(l)} \right)$. Vector quantization is performed using codebooks \(\mathcal{C}_H^{(l)}\) and \(\mathcal{C}_O^{(l)}\), mapping each latent vector to its nearest codeword, $\bar{Z}_H^{i,(l)} = \text{Quantize}\left( \hat{Z}_H^{i,(l)}, \mathcal{C}_H^{(l)} \right), \quad \bar{Z}_O^{(l)} = \text{Quantize}\left( \hat{Z}_O^{(l)}, \mathcal{C}_O^{(l)} \right)
$. The decoder reconstructs the inputs from the quantized latent representations, proceeding hierarchically. For human $i$, the decoder function is {\small$\hat{X}^i = D_H\left( \sum_{l=1}^L \bar{Z_{H}}^{i,(l)}; \phi_H \right)$}, where $D_H$ is a neural network with parameters $\phi_H$. Similarly, for the object, {\small$\hat{Y} = D_O\left(\sum_{l=1}^L \bar{Z}_O^{(l)}; \phi_O \right)$}. It is worth noting that our approach differs from other models like Priority-based VQ-VAE \cite{kong2023priority} and T2M-GPT\cite{zhang2023t2m} in terms of the latent representation used by the decoder. In our model, we aggregate the latent codes from all layers of the VQ-VAE before passing them to the decoder. This allows our decoder to work with non-discrete, continuous representations. As a result, our model can directly utilize the continuous representations generated by the diffusion model without the need for discrete mapping. In contrast, models like Priority-based VQ-VAE and T2M-GPT learn distributions over discrete latent codes, requiring an additional step to map the continuous diffusion outputs to discrete codes. Our training objective combines several loss terms, including reconstruction loss, commitment loss, codebook loss, alignment loss, hierarchical disentanglement loss, velocity smoothing loss, penetration loss, and contact loss.
The reconstruction loss measures the discrepancy between inputs and reconstructions, 
\small{
\begin{align*}
\mathcal{L}_{\text{recon}} = \sum_{i=1}^n \left| X^i - \hat{X}^i \right|_2^2 + \left| Y - \hat{Y} \right|_2^2
\end{align*}
} The commitment and codebook losses encourage alignment between encoder outputs and codebook embeddings,\newline
\small{
\vspace{-5mm}
\begin{align*}
\mathcal{L}_{\text{commit}}^{(l)} &= \sum_{i=1}^n \left| \hat{Z}_H^{i,(l)} - \text{sg}\left( \bar{Z}_H^{i,(l)} \right) \right|_2^2 + \left| \hat{Z}_O^{(l)} - \text{sg}\left( \bar{Z}_O^{(l)} \right) \right|_2^2, \\
\mathcal{L}_{\text{codebook}}^{(l)} &= \sum_{i=1}^n \left| \text{sg}\left( \hat{Z}_H^{i,(l)} \right) - \bar{Z}_H^{i,(l)} \right|_2^2 + \left| \text{sg}\left( \hat{Z}_O^{(l)} \right) - \bar{Z}_O^{(l)} \right|_2^2,
\end{align*}
}
where $\text{sg}(\cdot)$ denotes the stop-gradient operator. The alignment loss with description cues ensures latent representations align with semantic embeddings, \small{
$\mathcal{L}_{\text{align}}^{(l)} = \sum_{i=1}^n \left| \bar{Z}_H^{i,(l)} - \mathbf{e}_H^{(l)} \right|_2^2 + \left| \bar{Z}_O^{(l)} - \mathbf{e}_O^{(l)} \right|_2^2$
}.
The hierarchical disentanglement loss encourages different levels to capture distinct features:
\begin{equation*}
\mathcal{L}_{\text{disent}} = \sum_{l=1}^{L-1} \sum_{i=1}^n \left| \text{Cov}\left( \bar{Z}_H^{i,(l)}, \bar{Z}_H^{i,(l+1)} \right) \right|_F^2 + \left| \text{Cov}\left( \bar{Z}_O^{(l)}, \bar{Z}_O^{(l+1)} \right) \right|_F^2,
\end{equation*}
where $\text{Cov}(\cdot)$ denotes covariance, and $| \cdot |_F$ is the Frobenius norm. The penetration loss penalizes interpenetration between humans and objects, {\small
$\mathcal{L}_{\text{penetration}} = \sum_{i=1}^n \sum_{t=1}^T \max(0, -d(\hat{X}_t^i, \hat{Y}_t))$
}, where $d(\cdot, \cdot)$ computes the signed distance between human and object meshes.
The contact loss encourages plausible human-object contacts:
\small{
$\mathcal{L}_{\text{contact}} = \sum_{i=1}^n \sum_{t=1}^T \left| C(\hat{X}_t^i, \hat{Y}_t) - C(X_t^i, Y_t) \right|_2^2$,
}
where $C(\cdot, \cdot)$ computes the contact map between human and object meshes. The velocity smoothing loss encourages smooth motion transitions:
\small{
\begin{equation*}
\mathcal{L}_{\text{smooth}} = \sum_{i=1}^n \sum_{t=1}^{T-1} \left| \hat{X}_t^i - \hat{X}_{t-1}^i \right|_2^2 + \sum_{t=1}^{T-1} \left| \hat{Y}_t - \hat{Y}_{t-1} \right|_2^2.
\end{equation*}
}
The overall objective is:
\begin{align*}
\mathcal{L} =& \lambda_{\text{disent}} \mathcal{L}_{\text{disent}} + \lambda_{\text{smooth}} \mathcal{L}_{\text{smooth}}\\
&+ \lambda_{\text{penetration}} \mathcal{L}_{\text{penetration}} + \lambda_{\text{contact}} \mathcal{L}_{\text{contact}} + \mathcal{L}_{\text{recon}} \\
&+\sum_{l=1}^L \left( \lambda_{\text{commit}}^{(l)} \mathcal{L}_{\text{commit}}^{(l)} \lambda_{\text{codebook}}^{(l)} \mathcal{L}_{\text{codebook}}^{(l)} + \lambda_{\text{align}}^{(l)} \mathcal{L}_{\text{align}}^{(l)} \right)
\end{align*}
with weighting coefficients $\lambda$.
The hierarchical architecture enables the model to capture motion dynamics at multiple levels of abstraction, as described in Figure 1. The LLM provides hierarchical description cues for each level’s abstraction, guiding the model to associate latent representations with appropriate semantic concepts. Attention mechanisms across entities capture dependencies and interactions essential for understanding coordinated actions. The hierarchical disentanglement loss encourages different levels to focus on different features, preventing redundancy and promoting specialization, leading to more meaningful and interpretable representations. Codebooks are updated using exponential moving averages, and the straight-through estimator is employed to allow gradients to flow through quantization operations. Our hierarchical VQ-VAE with description cues effectively captures the multi-scale nature of human-object interactions, learning disentangled representations at different abstraction levels and aligning them with hierarchical semantic cues to enhance interpretability and facilitate advanced control in the motion latent space.
\subsection{Latent Diffusion with LLM Guidance}
Our objective is to generate realistic motion sequences involving multiple humans and objects, guided by hierarchical planning cues provided by a Large Language Model (LLM). To achieve this, we propose a denoising diffusion probabilistic model\cite{ho2020denoising,rasul2021autoregressive} that operates on the hierarchical latent codes learned by the VQ-VAE as described in Fig~\ref{fig:pipeline}. The model incorporates reasoning cues at multiple stages during the diffusion process, enabling it to generate motion sequences that align with the semantic intent expressed by the LLM.

Given a dataset $\mathcal{D} = \{(X_i, \mathbf{e}_i)\}_{i=1}^N$, where each motion sequence $X_i$ consists of the trajectories of $n$ humans $H = {H^1, \ldots, H^n}$ and $m$ objects $O = {O^1, \ldots, O^m}$ over $K$ time steps, and $\mathbf{e}_i = [\mathbf{e}_i^{(1)}, \ldots, \mathbf{e}_i^{(L)}]$ are the LLM-provided planning cues at $L$ reasoning steps, our diffusion model learns to generate motion sequences conditioned on these cues. To enhance the integration of planning cues \(\mathbf{E}_L = [\mathbf{e}_1, \ldots, \mathbf{e}_L]\) into the diffusion model, we associate each cue \(\mathbf{e}_l\) with relevant latent codes from the VQ-VAE codebook \(\mathcal{C}^{(l)}\) at level \(l\). After training the VQ-VAE, we compute associations between latent codes \(c \in \mathcal{C}^{(l)}\) and planning cues \(\mathbf{e}_l\) by learning embedding functions \(\phi_c^{(l)}(c)\) and \(\phi_e^{(l)}(\mathbf{e}_l)\) that map codes and cues into a shared semantic space. We optimize a contrastive loss to ensure associated pairs are close in the embedding space:
{\small
\begin{align*}
\mathcal{L}_{\text{assoc}}^{(l)} = - \sum_{(c, \mathbf{e}_l)} \log \frac{ \exp\left( \cos\left( \phi_c^{(l)}(c),\ \phi_e^{(l)}(\mathbf{e}_l) \right) / \tau \right) }{ \sum_{c' \in \mathcal{C}^{(l)}} \exp\left( \cos\left( \phi_c^{(l)}(c'),\ \phi_e^{(l)}(\mathbf{e}_l) \right) / \tau \right) },
\end{align*}
}

where \(\tau\) is a temperature parameter. For each planning cue \(\mathbf{e}_l\), we select the top \(u\) latent codes \(\{ c_{l,1}, \ldots, c_{l,u} \}\) most associated with \(\mathbf{e}_l\) based on cosine similarity. We augment the cue by concatenating the embeddings of these codes, $\tilde{\mathbf{e}}_l = \left[ \phi_e^{(l)}(\mathbf{e}_l);\ \phi_c^{(l)}(c_{l,1});\ \ldots;\ \phi_c^{(l)}(c_{l,u}) \right]$.

These augmented cues \(\tilde{\mathbf{E}}_L = [\tilde{\mathbf{e}}_1, \ldots, \tilde{\mathbf{e}}_L]\) are used in the denoising network, enhancing the model's capacity to generate motion sequences aligned with the planning cues by incorporating both semantic and structural information. This method leverages learned associations between planning cues and latent codes, improving the diffusion model's performance during the denoising process.

To prepare the input for the diffusion model, we aggregate the hierarchical latent codes from the VQ-VAE to form the initial latent representation $\mathbf{x}0$. This is achieved by summing the latent codes across all levels, $ \mathbf{x}_0 = \sum_{l=1}^L \left[ z^{(l)}_{H},\ z^{(l)}_{O} \right] \in \mathbb{R}^{F \times V \times K}$,  where $z^{(l)}_{H} \in \mathbb{R}^{F \times n \times K}$ are the latent codes for humans at level $l$, $z^{(l)}_{O} \in \mathbb{R}^{F \times m \times K}$ are object latent codes, $F$ is the feature dimension, $V = n + m$ is the total number of nodes, and $K$ is the number of time steps. This gives us a fully connected graph $\mathcal{G} = (\mathcal{V})$.

Our denoising network extends the U-Net architecture to handle spatio-temporal graph data. It incorporates downsampling and upsampling paths with residual connections, allowing the network to capture multi-scale temporal dependencies. We develop Motion Modeling Blocks (MM-Blocks) to process the data at different resolutions, effectively modeling the complex dynamics of motion sequences. Like previous diffusion-based models~\cite{rasul2021autoregressive}, we use positional encodings of the diffusion step $t \in {1, \ldots, T}$ and process it using a transformer positional embedding. This embedding, denoted as $\mathbf{p}_t$, is added between the temporal layers in each MM-Block, allowing the network to condition on the noise level and adapt its denoising strategy accordingly.

In the encoding path (downsampling), at each MM-Block $i$, we first apply Temporal Convolutional Networks (TCNs)~\cite{bai2018empirical} to capture temporal dependencies at multiple scales. The TCN at layer $i$ is defined as, $\mathbf{\overline{H}}_i = \text{TCN}(\mathbf{H}_{i-1}) + \mathbf{p}_t$, where $\mathbf{H}_{i-1} \in \mathbb{R}^{C_{i-1} \times V \times K_{i-1}}$ is the input from the previous layer, $C_{i-1}$ is the feature dimension, $K_{i-1}$ is the temporal length at layer $i-1$, and $\mathbf{p}_t \in \mathbb{R}^{C{i-1} \times 1 \times 1}$ is the positional embedding of the diffusion step $t$, broadcasted to match the dimensions. Adding the positional embedding allows the network to be aware of the current noise level, which is crucial for effective denoising, as observed by \cite{rasul2021autoregressive}. Next, we apply Graph Attention Networks (GATs)~\cite{velickovic2017graph} to model spatial dependencies, $\mathbf{H}_i = \text{GAT}(\mathbf{\overline{H}}_i, \mathbf{A})$, where $\mathbf{A}$ is the adjacency matrix denoting fully connected graph. The GAT allows the network to focus on important interactions between humans and objects by computing attention weights for the edges in the graph.

To incorporate the reasoning cues $\tilde{\mathbf{E}}_L = [\tilde{\mathbf{e}}_1, \ldots, \tilde{\mathbf{e}}_L]$, we perform cross-attention between the node features $\mathbf{H}_i$ and the reasoning cues at each MM-Block: \begin{equation} \mathbf{H}_i^{l} = \text{CrossAttn}(\mathbf{H}_i, \gamma_l(t) \cdot \tilde{\mathbf{E}}_L), \end{equation} for each $l \in {1, \ldots, L}$, where $\gamma_l(t)$ is a time-dependent modulation function. The time-dependent modulation function \(\gamma_l(t)\) dynamically adjusts the influence of each reasoning cue \(\tilde{\mathbf{e}}_L\) over the diffusion steps \(t\), emphasizing high-level planning cues at early steps and fine-grained details later. Specifically, for reasoning cue level \(l\), we define \(\gamma_l(t) = \lambda_l \exp(-k_l t / T)\) for high-level cues (\(l\) small), where \(\lambda_l\) is a scaling factor, \(k_l\) controls the rate of decay, and \(T\) is the total number of diffusion steps. This function decreases over time, giving high-level cues more influence when the data is noisy. For low-level cues(\(l\) large), we use \(\gamma_l(t) = \lambda_l [1 - \exp(-k_l t / T)]\), which increases over time, allowing fine-grained details to impact later steps when refining the motion. The rate of decay \(k_l\) can be a learnable parameter, enabling the model to adaptively determine the optimal influence schedule for each cue level. This modulation ensures the network focuses on appropriate aspects of the reasoning cues at each stage, effectively aligning the generated motion sequences with the LLM provided hierarchical plan. We then again pass the output through TCN and GAT layers. Finally, the outputs $\{\mathbf{H}_i^{l}\}_{l=1}^L$ are concatenated with $\mathbf{H}_i$ and passed to the next layer, ensuring that the semantic guidance is integrated throughout the network. We then perform downsampling to reduce the temporal dimension, $\mathbf{H}_i = \text{Downsample}(\mathbf{H}_i)$, which enables the network to capture long-range temporal dependencies. We repeat modelling spatial and temporal motion dynamics in alternate fashion, with downsampling steps.

In the decoding path (upsampling), we mirror the operations of the encoding path. Finally, we project the features back to the original latent space dimension to obtain the predicted noise: \begin{equation} \hat{\epsilon} = \text{Linear}(\mathbf{H}_0) \in \mathbb{R}^{F \times V \times K}. \end{equation}

The forward diffusion process\cite{ho2020denoising, rasul2021autoregressive,tashiro2021csdi} gradually adds Gaussian noise to the data,   $q(\mathbf{x}_t | \mathbf{x}_{t-1}) = \mathcal{N}(\mathbf{x}_t;\ \sqrt{1 - \beta_t} \mathbf{x}_{t-1},\ \beta_t \mathbf{I}),$ where ${\beta_t}_{t=1}^T$ is a predefined noise schedule. The reverse diffusion process aims to recover the original data from the noisy observations. The denoising network learns to predict the added noise $\epsilon$ at each diffusion step $t$, conditioned on the current noisy data $\mathbf{x}_t$, the graph structure $\mathcal{G}$, and the reasoning cues $\mathbf{E}_L$. The training objective is to minimize the expected L2 loss between the true noise $\epsilon$ and the predicted noise: \begin{equation} \mathcal{L}{\text{simple}} = \mathbb{E}{\mathbf{x}_0, \epsilon, t} \left[ \left| \epsilon - \epsilon_{\theta}(\mathbf{x}_t, t, \mathcal{G}, \mathbf{E}_L) \right|^2 \right], \end{equation} where $\mathbf{x}_t = \sqrt{\bar{\alpha}_t} \mathbf{x}_0 + \sqrt{1 - \bar{\alpha}_t}\ \epsilon$, with $\epsilon \sim \mathcal{N}(\mathbf{0}, \mathbf{I})$, and $\bar{\alpha}t = \prod{s=1}^t (1 - \beta_s)$. By minimizing this loss, the network learns to denoise the data effectively. Incorporating the reasoning cues $\mathbf{E}_L$ during diffusion allows the network to generate motion sequences that fulfill the intended actions and interactions. The time-dependent modulation function $\gamma_l(t)$ can be designed to emphasize high-level planning cues at early diffusion steps and fine-grained details at later steps, enabling the network to focus on different aspects of the reasoning at appropriate times.

During inference, we generate new motion sequences by starting from Gaussian noise $\mathbf{x}_T \sim \mathcal{N}(\mathbf{0}, \mathbf{I})$ and iteratively applying the denoising network: 
\begin{align*} \mathbf{x}_{t-1} = \frac{1}{\sqrt{1 - \beta_t}} \left( \mathbf{x}_t - \frac{\beta_t}{\sqrt{1 - \bar{\alpha}t}} \epsilon_{\theta}(\mathbf{x}_t, t, \mathcal{G}, \mathbf{E}_L) \right) + \sigma_t\ \mathbf{z}, \end{align*} 

where $\mathbf{z} \sim \mathcal{N}(\mathbf{0}, \mathbf{I})$ if $t > 1$, and $\sigma_t^2 = \beta_t$.

The combination of a U-shaped architecture with motion modeling blocks and hierarchical conditioning is particularly well-suited for generating motion sequences guided by LLM planning. The U-Net structure allows the network to capture temporal dependencies at multiple scales, essential for modeling complex motion dynamics over time. The use of GATs enables the network to model interactions between multiple humans and objects, capturing the spatial dependencies critical for realistic human-object interactions.

By integrating the LLM-based planning cues through cross-attention at every layer and modulating them over diffusion steps, the network generates motion sequences that are both semantically meaningful and physically plausible, aligning with the hierarchical planning provided by the LLM.

\begin{table*}[t!]
  \centering
  \resizebox{\textwidth}{!}{
  \rowcolors{2}{gray!10}{white} 
  \setlength{\tabcolsep}{4pt} 
  \renewcommand{\arraystretch}{1.2} 
  \begin{tabular}{@{}lccccccccccccccc@{}}
    \toprule
    \rowcolor{gray!30} 
    \multirow{2}{*}{\textbf{Methods(CORE-4D)}}  & \multicolumn{3}{c}{\textbf{R Precision}$\uparrow$} & \multirow{2}{*}{\textbf{FID}$\downarrow$} & \multirow{2}{*}{\textbf{MM Dist}$\downarrow$}  & \multirow{2}{*}{\textbf{Diversity}$\rightarrow$} & \multirow{2}{*}{\textbf{MModality}$\uparrow$} &
    \multirow{2}{*}{\textbf{Methods}\textbf{(InterHuman)}}  & \multicolumn{3}{c}{\textbf{R Precision}$\uparrow$} & \multirow{2}{*}{\textbf{FID}$\downarrow$} & \multirow{2}{*}{\textbf{MM Dist}$\downarrow$}  & \multirow{2}{*}{\textbf{Diversity}$\rightarrow$} & \multirow{2}{*}{\textbf{MModality}$\uparrow$}\\
    \cmidrule(lr){2-4} \cmidrule(lr){10-12} 
     & \textbf{Top 1} & \textbf{Top 2}  & \textbf{Top 3} & & & & & & \textbf{Top 1} & \textbf{Top 2}  & \textbf{Top 3} \\
    \midrule
                Real & $0.312^{\pm0.007}$ & $0.587^{\pm0.006}$ & $0.673^{\pm0.006}$ & $0.005^{\pm0.0005}$ & $4.124^{\pm0.019}$ & $8.151^{\pm0.091}$ & - & Real & $0.452^{\pm0.008}$ & $0.610^{\pm0.009}$ & $0.701^{\pm0.008}$ & $0.273^{\pm0.007}$ & $3.755^{\pm0.008}$ & $7.948^{\pm0.064}$ & {-}\\
      \midrule
      TEMOS~\cite{petrovich22temos} & $0.065^{\pm0.006}$ & $0.179^{\pm0.006}$ & $0.211^{\pm0.005}$ & $9.214^{\pm0.0758}$ & $8.536^{\pm0.019}$ & $4.671^{\pm0.091}$ & $0.510^{\pm0.052}$ & TEMOS~\cite{petrovich22temos} & $0.224^{\pm0.010}$ & $0.316^{\pm0.013}$ & $0.450^{\pm0.018}$ & $17.375^{\pm0.043}$ & $6.342^{\pm0.015}$ & $6.939^{\pm0.071}$ & $0.535^{\pm0.014}$ \\ 
      T2M~\cite{humanml3d} & $0.195^{\pm0.003}$ & $0.141^{\pm0.003}$ & $0.267^{\pm0.002}$ & $11.258^{\pm0.0694}$ & $5.867^{\pm0.013}$ & $2.738^{\pm0.078}$ & $1.672^{\pm0.041}$  & T2M~\cite{humanml3d} & $0.238^{\pm0.012}$ & $0.325^{\pm0.012}$ & $0.464^{\pm0.014}$ & $13.769^{\pm0.072}$ & $5.731^{\pm0.018}$ & $7.046^{\pm0.022}$ & $1.387^{\pm0.076}$\\

      MDM~\cite{mdm} &  $0.163^{\pm0.013}$ & $0.257^{\pm0.010}$ & $0.348^{\pm0.008}$ & $9.671^{\pm0.0629}$ & $10.219^{\pm0.020}$ & $\mathbf{7.395^{\pm0.090}}$ & $\mathbf{3.526^{\pm0.074}}$ & MDM~\cite{mdm} & $0.153^{\pm0.009}$ & $0.260^{\pm0.011}$ & $0.339^{\pm0.012}$ & $9.167^{\pm0.056}$ & $7.125^{\pm0.018}$ & $\mathbf{7.602^{\pm0.045}}$ & $2.355^{\pm0.080}$\\
      MDM(GRU)~\cite{mdm} &  $0.168^{\pm0.009}$ & $0.279^{\pm0.008}$ & $0.361^{\pm0.010}$ & $9.587^{\pm0.1382}$ & $10.228^{\pm0.025}$ & $6.951^{\pm0.151}$ & $3.170^{\pm0.046}$ & MDM(GRU)~\cite{mdm} &  $0.179^{\pm0.006}$ & $0.299^{\pm0.005}$ & $0.387^{\pm0.007}$ & $32.617^{\pm0.1221}$ & $9.557^{\pm0.019}$ & $7.003^{\pm0.134}$ & $\mathbf{3.430^{\pm0.035}}$\\
      ComMDM~\cite{commdm} & $0.187^{\pm0.005}$ & $0.256^{\pm0.007}$ & $0.301^{\pm0.007}$ & $9.217^{\pm0.0727}$ & $7.541^{\pm0.023}$ & $5.367^{\pm0.080}$ & $0.721^{\pm0.065}$ & ComMDM~\cite{commdm} & $0.223^{\pm0.010}$ & $0.334^{\pm0.008}$ & $0.466^{\pm0.012}$ & $7.069^{\pm0.054}$ & $6.212^{\pm0.021}$ & $7.244^{\pm0.038}$ & $1.822^{\pm0.052}$\\
      InterGen~\cite{liang2023intergen} & $0.206^{\pm0.007}$ & $0.312^{\pm0.008}$ & $0.401^{\pm0.008}$ & $7.217^{\pm0.2321}$ & $10.251^{\pm0.017}$ & $6.162^{\pm0.225}$ & $3.402^{\pm0.063}$ & InterGen~\cite{liang2023intergen} & $0.371^{\pm0.010}$ & $0.515^{\pm0.012}$ & $0.624^{\pm0.010}$ & $5.918^{\pm0.079}$ & $5.108^{\pm0.014}$ & $7.387^{\pm0.029}$ & $2.141^{\pm0.063}$\\
      \midrule
      \textbf{\ourmodel} & $\mathbf{0.229^{\pm0.008}}$ & $\mathbf{0.332^{\pm0.009}}$ & $\mathbf{0.435^{\pm0.009}}$ & $\mathbf{6.890^{\pm0.2198}}$ & $\mathbf{5.526^{\pm0.016}}$ & ${7.373^{\pm0.237}}$ & ${3.589^{\pm0.066}}$ & \textbf{\ourmodel} & $\mathbf{0.383^{\pm0.005}}$ & $\mathbf{0.547^{\pm0.006}}$ & $\mathbf{0.657^{\pm0.006}}$ & $\mathbf{4.987^{\pm0.2061}}$ & $\mathbf{4.992^{\pm0.012}}$ & ${7.515^{\pm0.214}}$ & ${2.872^{\pm0.057}}$\\
      \midrule
      {55 Sampling steps} & $0.227^{\pm0.008}$ & $0.330^{\pm0.008}$ & $0.433^{\pm0.008}$ & $6.923^{\pm0.2201}$ & $5.546^{\pm0.017}$ & $7.152^{\pm0.235}$ & $3.561^{\pm0.065}$ & {55 Sampling steps} & $0.380^{\pm0.006}$ & $0.545^{\pm0.007}$ & $0.655^{\pm0.006}$ & $5.016^{\pm0.2070}$ & $5.001^{\pm0.013}$ & $7.458^{\pm0.212}$ & $2.850^{\pm0.058}$\\
      {15 Sampling steps} & $0.215^{\pm0.007}$ & $0.322^{\pm0.008}$ & $0.429^{\pm0.008}$ & $7.030^{\pm0.2101}$ & $5.865^{\pm0.018}$ & $6.712^{\pm0.228}$ & $3.515^{\pm0.065}$ & {15 Sampling steps} & $0.376^{\pm0.006}$ & $0.539^{\pm0.007}$ & $0.650^{\pm0.006}$ & $5.452^{\pm0.2053}$ & $5.215^{\pm0.013}$ & $7.381^{\pm0.218}$ & $2.759^{\pm0.060}$\\
      {5 Sampling steps} & $0.188^{\pm0.007}$ & $0.265^{\pm0.008}$ & $0.338^{\pm0.008}$ & $8.150^{\pm0.2365}$ & $7.198^{\pm0.019}$ & $6.135^{\pm0.221}$ & $2.876^{\pm0.062}$ & {5 Sampling steps} & $0.258^{\pm0.007}$ & $0.408^{\pm0.008}$ & $0.520^{\pm0.007}$ & $8.112^{\pm0.082}$ & $5.901^{\pm0.014}$ & $7.045^{\pm0.031}$ & $1.835^{\pm0.062}$\\
     
      w/o Hierarchy & $0.201^{\pm0.007}$ & $0.309^{\pm0.008}$ & $0.411^{\pm0.008}$ & $7.452^{\pm0.2381}$ & $5.582^{\pm0.018}$ & $6.995^{\pm0.224}$ & $3.209^{\pm0.058}$ & w/o Hierarchy & $0.355^{\pm0.009}$ & $0.521^{\pm0.010}$ & $0.632^{\pm0.009}$ & $5.543^{\pm0.2154}$ & $5.048^{\pm0.015}$ & $7.137^{\pm0.201}$ & $2.492^{\pm0.061}$\\
        w/o LLM & $0.208^{\pm0.007}$ & $0.315^{\pm0.008}$ & $0.419^{\pm0.008}$ & $7.235^{\pm0.2305}$ & $5.561^{\pm0.017}$ & $6.549^{\pm0.230}$ & $3.152^{\pm0.063}$ & w/o LLM & $0.362^{\pm0.008}$ & $0.528^{\pm0.009}$ & $0.639^{\pm0.008}$ & $5.326^{\pm0.2087}$ & $5.027^{\pm0.014}$ & $6.691^{\pm0.207}$ & $2.435^{\pm0.059}$\\
 w/o Time Modulation & $0.218^{\pm0.008}$ & $0.317^{\pm0.009}$ & $0.420^{\pm0.009}$ & $7.071^{\pm0.2251}$ & $5.556^{\pm0.017}$ & $7.263^{\pm0.234}$ & $3.474^{\pm0.065}$ & w/o Time Modulation & $0.372^{\pm0.007}$ & $0.536^{\pm0.008}$ & $0.647^{\pm0.007}$ & $5.162^{\pm0.2129}$ & $5.021^{\pm0.013}$ & $7.405^{\pm0.211}$ & $2.767^{\pm0.062}$\\
      \hline
    \bottomrule
  \end{tabular}} 
\caption{Experimental results and Ablation studies for text-conditioned interaction generation on the CORE-4D and InterHuman datasets, where $\pm$ indicates 95\% confidence interval and $\rightarrow$ means the closer the better. \textbf{Bold} indicates best results.}
\vspace{-7mm}
  \label{tab:merged_results}
\end{table*}

\begin{table}[tb]
  \centering
  \resizebox{0.47\textwidth}{!}{
  \footnotesize
  \begin{tabular}{c|c|c|c|c|c}
    \hline
    Test Set & Method & $RR.J_e$ (mm, $\downarrow$) & $RR.V_e$ (mm, $\downarrow$) & $C_{\text{acc}}$ ($\%$, $\uparrow$) & $FID$ ($\downarrow$) \\
    \hline
    \multirow{3}{*}{S1} & MDM~\cite{mdm} & 138.0 ($\pm$ 0.3) & 194.6 ($\pm$ 0.2) & 76.9 ($\pm$ 0.5) & 7.7 ($\pm$ 0.2) \\
    \cline{2-6}
    & OMOMO~\cite{zhang2024core4d} & 137.8 ($\pm$ 0.2) & 196.7 ($\pm$ 0.3) & 78.2 ($\pm$ 0.5) & 8.3 ($\pm$ 0.6) \\
    \cline{2-6}
    & \ourmodel & \textbf{131.2 ($\pm$ 0.2)} & \textbf{185.1 ($\pm$ 0.2)} & \textbf{80.5 ($\pm$ 0.4)} & \textbf{7.2 ($\pm$ 0.2)} \\
    \hline
    \multirow{3}{*}{S2} & MDM~\cite{mdm} & 145.9 ($\pm$ 0.2) & 208.2 ($\pm$ 0.2) & 76.7 ($\pm$ 0.1) & 7.7 ($\pm$ 0.2) \\
    \cline{2-6}
    & OMOMO~\cite{zhang2024core4d} & 145.2 ($\pm$ 0.6) & 209.9 ($\pm$ 1.0) & 77.8 ($\pm$ 0.3) & 8.3 ($\pm$ 1.0) \\
    \cline{2-6}
    & \ourmodel & \textbf{138.5 ($\pm$ 0.5)} & \textbf{198.7 ($\pm$ 0.8)} & \textbf{79.9 ($\pm$ 0.2)} & \textbf{7.3 ($\pm$ 0.8)} \\
    \hline
  \end{tabular}}
  \caption{\small Quantitative results on object-conditioned interaction synthesis on CORE-4D.\vspace{-8mm}}
  \label{tab:results_synthesis}
\end{table}
\section{Experimentation and Results}
\label{sec:experiments}

\paragraph{\textbf{Implementation Details}} Our model consists of a hierarchical VQ-VAE with $L=6$ levels, each with a codebook size of $512 \times 512$ (latent dimension 512) and two Conv1D blocks per level per entity (kernel size 3, residual connections), similar to T2M-GPT~\cite{zhang2023t2m}.  Vector quantization is performed using the straight-through estimator, and hierarchical planning cues are generated via GPT-4~\cite{chatgpt} and embedded using CLIP ViT-B/32~\cite{radford2021learning}, associated with the VQ-VAE codebooks through contrastive learning (temperature $\tau = 0.07$, top $u=8$ latent codes per level). The latent diffusion model is based on a U-Net architecture with $M=4$ Motion Modeling Blocks (MM-Blocks), each consisting of Temporal Convolutional Networks (TCNs) with kernel sizes $\{3, 5, 7\}$~\cite{bai2018empirical} and Graph Attention Networks (GATs) with 8 attention heads~\cite{velickovic2017graph}, capturing spatio-temporal dependencies.For training, we use the Adam optimizer~\cite{kingma2020method} for VQ-VAE with a learning rate of $1 \times 10^{-4}$ and AdamW~\cite{loshchilov2017decoupled} for the diffusion model with a learning rate of $2 \times 10^{-4}$, both with cosine annealing, gradient clipping (max norm 1.0), and weight decay of $1 \times 10^{-5}$. For CORE-4D~\cite{zhang2024core4d}, we train for 50K iterations with a learning rate of $2 \times 10^{-4}$ and an additional 30K iterations with a reduced learning rate of $1 \times 10^{-5}$. For InterHuman~\cite{liang2023intergen}, we train for 200K iterations at $2 \times 10^{-4}$ and 100K iterations at $1 \times 10^{-5}$. We use a batch size of 256 for both datasets and apply the Adam optimizer with $[\beta_1, \beta_2] = [0.9, 0.99]$ and an exponential moving constant $\lambda=0.99$. Loss terms include $\lambda_{\text{recon}} = 1.0$, $\lambda_{\text{commit}}^{(l)} = 0.25$ per level, $\lambda_{\text{codebook}}^{(l)} = 0.25$ per level, $\lambda_{\text{align}}^{(l)} = 0.5$ per level, $\lambda_{\text{smooth}} = 0.1$\cite{zhang2023motiongpt}, $\lambda_{\text{penetration}} = 10.0$\cite{li2023task}, and $\lambda_{\text{contact}} = 5.0$. The hierarchical disentanglement loss is weighted by $\lambda_{\text{disent}} = 1.0$. The diffusion model uses 1000 diffusion steps and we test for 5, 15, 55, 100 DDIM~\cite{ddim} sampling steps during inference. The hierarchical cue modulation function applies exponential decay for high-level cues and increasing influence for low-level cues across diffusion steps

We train \ourmodel on the CORE-4D dataset~\cite{zhang2024core4d}, which contains 998 motion sequences of human-object-human interactions spanning 5 object categories. We annotates the motion sequences with textual descriptions, the annotated text-motion dataset has an average length of 8.54 words, totaling 8,542 words. We split the dataset into training, validation, and test sets with a ratio of 0.8, 0.05, and 0.15, respectively. We also evaluate our model on the InterHuman dataset~\cite{liang2023intergen} for multi-human generation, which includes 6,022 motions with 16,756 unique descriptions. We use same train/test formulation as ~\cite{liang2023intergen}. We additionally also train our model for single human motion generation on KIT-ML\cite{plappert2016kit} and HumanML3D\cite{9880214} Dataset, the visualisations and comparisons are available in attached video. 
\paragraph{\textbf{Evaluation Metrics}}
For text-conditioned generation on CORE-4D and InterHuman, we adopt the metrics from InterGen~\cite{liang2023intergen}: (1) \textit{FID}, (2) \textit{R-Precision}, (3) \textit{Diversity}, (4) \textit{Multimodality (MModality)}, and (5) \textit{MM Dist}. For additional tasks on CORE-4D, we follow their own metrics~\cite{zhang2024core4d}: (1) \textit{RR.J\textsubscript{e}}, (2) \textit{RR.V\textsubscript{e}}, and (3) \textit{C\textsubscript{acc}}. All evaluations are run 20 times (except MModality, 5 times) with average results reported with a 95\% confidence interval. For detailed descriptions of these metrics, we refer readers to \cite{liang2023intergen,zhang2024core4d}.

\paragraph{\textbf{Baselines}} For text-conditioned generation on the CORE-4D dataset, we compare \ourmodel against state-of-the-art methods, including TEMOS~\cite{petrovich22temos}, T2M~\cite{humanml3d}, MDM~\cite{mdm}, MDM-GRU~\cite{mdm,cho2014learning}, ComMDM~\cite{commdm}, and InterGen~\cite{liang2023intergen}. We modify these models to handle two-person interactions and train them on the CORE-4D dataset. For the additional tasks on the CORE-4D dataset, we compare against MDM~\cite{mdm}, a one-stage motion diffusion model, and OMOMO~\cite{zhang2024core4d}, a two-stage approach for object-conditioned human motion generation.

\subsection{Results}

\subsubsection{\textbf{Text-Conditioned Generation}}
\paragraph{Results on CORE-4D} Table~\ref{tab:merged_results} (left) presents the results of text-conditioned generation on the CORE-4D dataset. \ourmodel outperforms all baselines across most metrics, achieving the highest R-Precision scores, lowest FID, and best diversity. The hierarchical VQ-VAE effectively captures multi-scale motion dynamics, while the LLM-guided diffusion model generates motions that align well with the textual descriptions. The incorporation of hierarchical planning cues enables \ourmodel to generate more coherent and diverse interactions compared to the baselines.
\paragraph{Results on InterHuman} We further evaluate \ourmodel on the InterHuman dataset for multi-human generation. Table~\ref{tab:merged_results} (right) shows the comparison with state-of-the-art methods. \ourmodel achieves superior performance across nearly all metrics, demonstrating its effectiveness in generating diverse and realistic multi-human interactions. The hierarchical modeling of motion dynamics and the incorporation of LLM-guided planning enable \ourmodel to better capture the complexities of human-human interactions compared to the baselines.
\subsubsection{Object-Conditioned Generation on CORE-4D}
We evaluate \ourmodel on the task of object-conditioned human motion generation on the CORE-4D dataset. Given an object geometry sequence, the goal is to generate two-person collaboration motions using the SMPL-X model~\cite{SMPLX}. Table~\ref{tab:results_synthesis} presents the quantitative results, comparing \ourmodel with MDM~\cite{mdm} and OMOMO~\cite{zhang2024core4d}. \ourmodel achieves the lowest joint and vertex position errors, highest contact accuracy , and best motion quality (FID) on both test sets (S1 and S2). The hierarchical modeling and LLM guidance enable \ourmodel to generate more precise and realistic human-object interactions compared to the baselines.
\subsubsection{Ablation Studies}
\begin{figure}
\centering
\includegraphics[width=\linewidth]{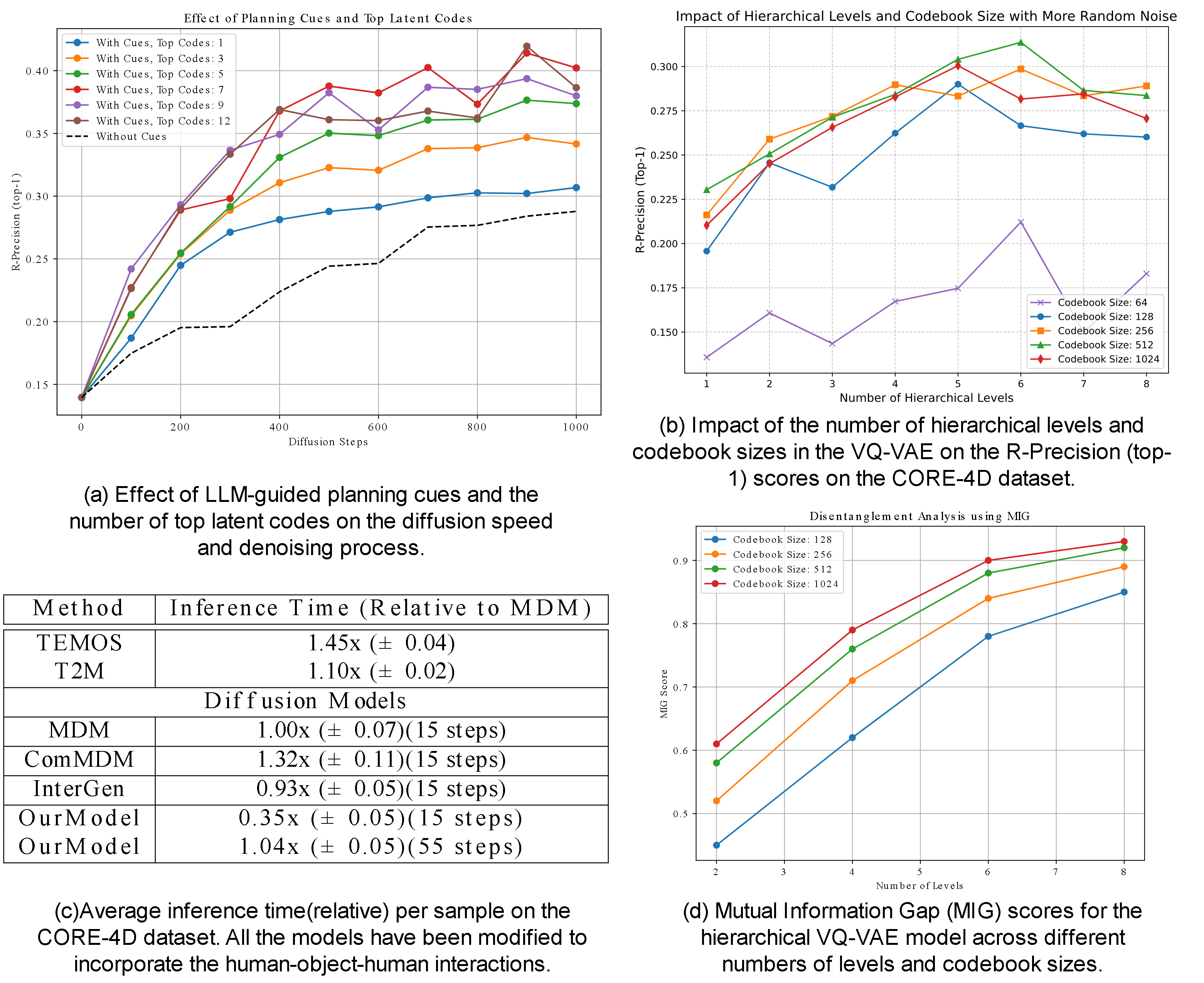}
\vspace{-7mm}
\caption{Ablation Studies}
\label{fig:ablation}
\vspace{-6mm}
\end{figure}

We conduct ablation studies on the CORE-4D dataset to validate the effectiveness of the proposed components in \ourmodel. Table~\ref{tab:merged_results} (bottom) presents the results. Removing the hierarchical structure in the VQ-VAE (w/o Hierarchy) significantly drops performance across all metrics, highlighting the importance of modeling motion dynamics at multiple scales. Removing LLM guidance (w/o LLM) also decreases performance, demonstrating the effectiveness of incorporating hierarchical planning cues. Replacing time-dependent modulation with a fixed weighting scheme (w/o Time Modulation) degrades performance, indicating the benefit of adaptively adjusting the influence of planning cues over diffusion steps. These studies confirm that the hierarchical VQ-VAE, LLM guidance, and time-dependent modulation are essential components of \ourmodel, contributing to its superior performance in generating collaborative human-object-human interactions.
\paragraph{Impact of Hierarchical Levels and Codebook Size}
We evaluate \ourmodel's performance with different numbers of hierarchical levels and codebook sizes in the VQ-VAE architecture. Figure~\ref{fig:ablation} shows the R-Precision (top-1) scores on the CORE-4D dataset as we vary the number of levels from 1 to 8 and the codebook sizes from 128 to 1024. Increasing the number of levels initially improves performance, with the best results achieved at 6 levels for all codebook sizes, indicating that the hierarchical structure effectively captures motion dynamics at multiple scales. However, further increasing levels beyond 6 slightly degrades performance, likely due to overfitting and increased model complexity. We find that codebook sizes of 256 and 512 strike a good balance between expressiveness and efficiency, with 512 yielding the best overall performance across different levels.
\paragraph{Effect of Planning Cues and Top Latent Codes}
We analyze the effect of LLM-guided planning cues and the number of top latent codes on diffusion speed, i.e., the number of steps required to generate high-quality motions. Figure~\ref{fig:ablation} compares the R-Precision (top-1) scores of \ourmodel with and without planning cues at different diffusion steps and the impact of using different numbers of top latent codes. Incorporating planning cues significantly accelerates the diffusion process, allowing the model to generate high-quality motions in fewer steps. Increasing the number of top latent codes further improves performance, with diminishing returns beyond 7-8 codes, demonstrating the effectiveness of cues and top latent codes in guiding the denoising process and highlighting the efficiency gains achieved by our approach.
\paragraph{Impact of Hierarchical Structure and Codebook Size on Latent Disentanglement}
Our ablation study investigates the influence of the hierarchical structure and codebook size on the disentanglement of latent representations in our VQ-VAE model. Figure~\ref{fig:ablation} reveals that increasing the number of levels consistently improves disentanglement, as evidenced by higher MIG\cite{chen2018isolating} scores, suggesting that the hierarchical structure effectively captures disentangled representations at different levels of abstraction, with higher levels focusing on more abstract patterns. Similarly, larger codebook sizes lead to better disentanglement for a given number of levels, indicating that a larger codebook enables more expressive and disentangled representations. However, the diminishing gaps between the lines at higher levels and codebook sizes imply that the benefits of increasing these hyperparameters saturate beyond certain thresholds, highlighting the importance of balancing model complexity and computational efficiency when designing hierarchical VQ-VAE architectures for learning disentangled representations of human motion data.
These ablation studies provide insights into the functioning and effectiveness of \ourmodel. The evaluation of different hierarchical levels and codebook sizes highlights the importance of finding the optimal balance between model complexity and expressiveness. The latent space disentanglement analysis demonstrates the hierarchical VQ-VAE's ability to capture distinct and independent features at different levels of abstraction. The analysis of diffusion speed showcases the efficiency gains achieved by incorporating planning cues and utilizing top latent codes, enabling faster generation of high-quality motions.
\subsection{Qualitative Analysis}
Attached video presents qualitative examples of generated collaborative human-object-human interactions by \ourmodel and the baselines on the CORE-4D dataset. \ourmodel generates more realistic and coherent interactions compared to the baselines, accurately capturing the coordination between the two humans and their interactions with the object. The generated motions align well with the input text descriptions, demonstrating the effectiveness of the LLM-guided planning cues in controlling the generation process. In contrast, the baselines struggle to generate precise and coordinated interactions, often resulting in unrealistic or inconsistent motions.

\paragraph{Runtime Analysis}
We compare the inference time of \ourmodel with the baselines on the CORE-4D dataset. Table~\ref{fig:ablation} presents the average relative inference time per sample for each method with respect to the MDM~\cite{tevet2022human} runtime. \ourmodel achieves significantly faster inference. Furthermore, we tested our performance for different DDIM sampling steps (5, 15, 55, and 100 in the main model). As expected, with an increase in the number of steps, the generation quality improves. However, the improvement in generation quality from 55 to 100 steps is minor, while the generation time nearly doubles. Notably, we observe that with just 15 steps, we achieve relatively better generation quality than InterGen~\cite{liang2023intergen}, and our model with 15 steps is faster than the MDM model (15 DDIM steps) by 65\%. Additionally, our near-best performance (55 DDIM steps) has a runtime similar to MDM. Thus, our hierarchical VQ-VAE enables efficient compression and decompression, while the LLM-guided cues and codebook associations provide curated motion priors, allowing the diffusion model to denoise in fewer steps and generate smoother motion faster. In contrast, other baselines require longer inference times due to their complex architectures and the need to generate motions in the original high-dimensional space.
\section{Discussion and Limitations}
The experimental results demonstrate the effectiveness of \ourmodel in generating realistic and diverse collaborative human-object-human interactions. The hierarchical VQ-VAE architecture captures motion dynamics at multiple scales, while the LLM-guided diffusion model generates motions that align well with textual descriptions and planning cues. The incorporation of hierarchical planning cues from the LLM allows for more coherent and controllable generation, as evidenced by \ourmodel's superior performance across various metrics and datasets.

However, there are some limitations to our approach. First, \ourmodel does not explicitly model the physical interactions between humans and objects, relying on learned motion priors to generate plausible interactions. Incorporating explicit physics modeling could further improve the realism and consistency of the generated motions. Second, the current approach generates motions from scratch based on input text and object geometry, but does not allow for fine-grained editing or control over specific aspects of the motion. Extending the model to support motion editing and user-guided refinement could enhance its practical utility. \\
\textbf{Despite} these limitations, \ourmodel represents a significant step towards generating realistic and diverse collaborative human-object-human interactions. Our approach can be seamlessly extended to generate collaborative interactions between humanoid robots and objects. By training our model on motion data from humanoid robots, we can generate realistic and diverse interactions that mimic human-like behaviors. This extension has significant implications for the deployment of humanoid robots in various real-world scenarios, where they are expected to collaborate with objects and other agents in a human-like manner.\\ \textbf{The} proposed approach opens up new possibilities for applications in robotics, virtual reality, and computer graphics, where generating plausible and coordinated multi-agent interactions is crucial. Future work could explore incorporating explicit physics modeling, supporting motion editing and user control, and extending the approach to handle a larger variety of objects and interaction scenarios.
\section*{Acknowledgement}
This material is based upon work supported in part by the DEVCOM Army Research Laboratory under cooperative agreement W911NF2020221.

\hypersetup{breaklinks=true}
\bibliographystyle{IEEEtran}
\bibliography{references}

\end{document}